\documentclass[10pt,twocolumn,letterpaper]{article}

 \usepackage[pagenumbers]{cvpr} 

\usepackage{graphicx}
\usepackage{amsmath}
\usepackage{amssymb}
\usepackage{booktabs}
\usepackage{algpseudocode}
\usepackage{multirow}

\usepackage{tabularx}

\usepackage{caption}

\usepackage{array}
\usepackage{ragged2e}
\usepackage{makecell}
\usepackage{multirow}
\usepackage{tabularx}

\definecolor{cvprblue}{rgb}{0.21,0.49,0.74}
\usepackage[pagebackref,breaklinks,colorlinks,allcolors=cvprblue]{hyperref}

\def\paperID{*****} 
\def\confName{CVPR}
\def\confYear{2026}

\title{RL‑ScanIQA: Reinforcement-Learned Scanpaths \\ for Blind 360° Image Quality Assessment}

\author{Yujia~Wang\textsuperscript{\rm 1},
 Yuyan~Li\textsuperscript{\rm 2},
 Jiuming~Liu\textsuperscript{\rm 3}, Fang-Lue~Zhang\textsuperscript{\rm 4}\thanks{Corresponding Author.}\\
 Xinhu~Zheng\textsuperscript{\rm 2},
 Neil.A~Dodgson\textsuperscript{\rm 1}\\
  {\textsuperscript{\rm 1}Victoria University of Wellington} \ \ 
  \textsuperscript{\rm 2}The Hong Kong University of Science and Technology (Guangzhou)\\
  {\textsuperscript{\rm 3}University of Cambridge} \ \ \ 
  {\textsuperscript{\rm 4}University of New South Wales}\\
}

\begin{document}
\maketitle
\begin{abstract}

Blind 360° image quality assessment (IQA) aims to predict perceptual quality for panoramic images without a pristine reference. Unlike conventional planar images, 360° content in immersive environments restricts viewers to a limited viewport at any moment, making viewing behaviors critical to quality perception. Although existing scanpath-based approaches have attempted to model viewing behaviors by approximating the human view‑then‑rate paradigm, they treat scanpath generation and quality assessment as separate steps, preventing end-to-end optimization and task-aligned exploration. To address this limitation, we propose RL‑ScanIQA, a reinforcement‑learned framework for blind 360° IQA. RL-ScanIQA optimize a PPO-trained scanpath policy and a quality assessor, where the policy receives quality-driven feedback to learn task-relevant viewing strategies. To improve training stability and prevent mode collapse, we design multi-level rewards, including scanpath diversity and equator-biased priors. We further boost cross‑dataset robustness using distortion‑space augmentation together with rank‑consistent losses that preserve intra‑image and inter‑image quality orderings. Extensive experiments on three benchmarks show that RL‑ScanIQA achieves superior in‑dataset performance and cross‑dataset generalization.

\end{abstract}    
\section{Introduction}

Image quality assessment (IQA) aims to predict the human perceptual quality of images \cite{ding2020image,gu2020giqa}. Due to the limited availability of compared reference images in real-world scenarios, blind (no-reference) IQA (BIQA) has recently gained substantial research attention \cite{mittal2012no,shin2024blind,zhou2025blind}. In contrast to full-reference methods \cite{sheikh2006statistical}, BIQA supports more downstream tasks such as image restoration \cite{liang2021swinir}, autonomous driving \cite{liu2023regformer,liu2024difflow3d,ni2025recondreamer}, and AI-generated content assessment \cite{liu2024ntire,ni2025wonderfree,liu20254dstr,ma2026gor,li2025one}.

The widespread adoption of virtual and augmented reality (VR/AR) has driven extensive use of 360° omnidirectional images \cite{fang2022perceptual,JCST-2303-13210}. Assessing and guaranteeing the visual quality of panoramic videos remains an open challenge \cite{fan2024learned}. Unlike planar images, 360° content is captured on a sphere and experienced through sequential head and eye movements within immersive environments \cite{sitzmann2018saliency,fang2022perceptual,scanTD}. Because viewers perceive only a limited viewport at any moment, diverse scanpaths determine which distortions are encountered and thus perceived quality depends not only on the content but also on the viewer’s visual attentional trajectory \cite{sun2017cviqd,duan2018perceptual,wu2023assessor360,fan2024learned,sui2025perceptual}. These considerations demand the development of novel BIQA models that effectively capture the spherical geometry inherently embedded in panoramic images and also users' viewing behaviors.


\begin{figure}
\includegraphics[width=\linewidth]{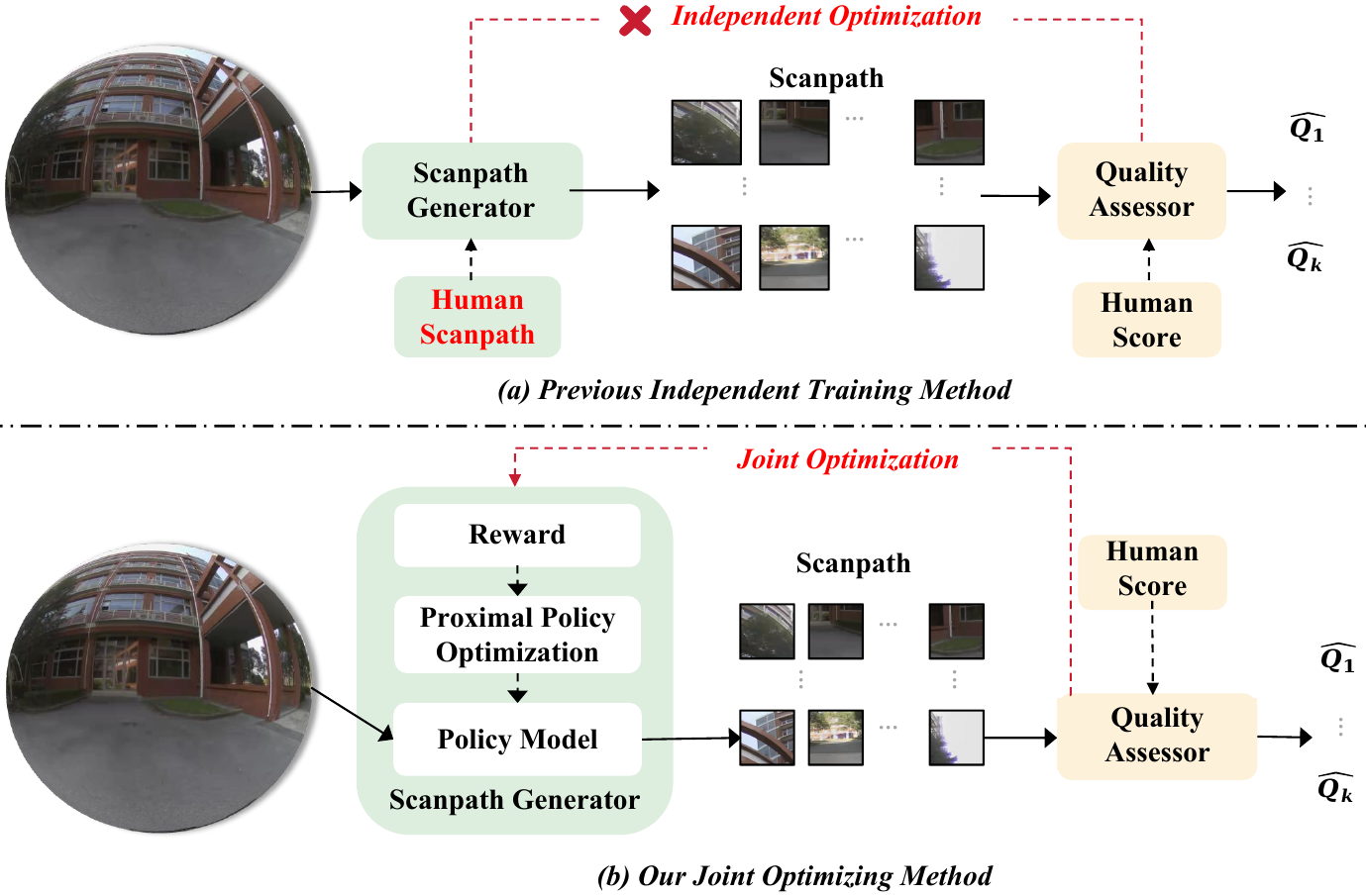}
\vspace{-7mm}
  \caption{Comparison of previous works and ours. \textit{a}: Previous works train the scanpath generator and quality assessor independently, typically with human scanpath supervision. \textit{b}: Our method requires no human scanpath and jointly optimizes these two modules end-to-end with a reinforcement learning policy.
}
  \label{introduction}
  \vspace{-4mm}
\end{figure}


Existing BIQA methods for 360° images typically fall into three categories: (1) direct, full-image analysis based on equirectangular projection (ERP) \cite{mittal2012making, mittal2012no}; (2) viewport sampling with predefined or static strategies \cite{li2019viewport,azevedo2020viewport}; and (3) scanpath-based exploration guided by observed user behavior \cite{wu2023assessor360,sui2025perceptual,fan2024learned}. ERP-based models, which treat the 360° image as a planar projection, suffer from inherent spatial biases and distortions. Viewport-based methods, such as patch-wise CNNs, often neglect the sequential nature of user exploration by relying on fixed sampling locations. Recent scanpath-based approaches offer improved approximations of human perception by simulating realistic viewing patterns. However, a key limitation still remains: These pipelines generally treat scanpath generation as a separate, externally trained pre-processing step decoupled from the quality prediction task, and are typically trained using heuristics or independent supervision to imitate human behaviors \cite{wu2023assessor360,sui2025perceptual}.

To address this problem, we propose RL-ScanIQA, an end-to-end, task-driven framework for blind 360° image quality assessment that jointly learns viewport sampling (scanpath) and quality assessment. As shown in Figure~\ref{introduction}, RL-ScanIQA has two jointly optimized modules, where scanpath generation is first formulated as a reinforcement learning (RL) problem in which the agent receives feedback directly from the subsequent IQA assessor, enabling the RL policy to learn task-relevant exploration trajectories from quality-prediction feedback instead of merely imitating human gaze. Unlike prior approaches that treat scanpath generation as a separate step, we formulate it as a sequential decision-making process, directly aligned with the IQA objective. Conceptually, RL-ScanIQA reformulates blind 360° IQA as active perception with end-to-end task-driven scanpath learning, rather than in any individual reward term or augmentation taken in isolation. Furthermore, distortion-space data augmentation with rank-consistency constraints is employed to improve generalization capability under cross-dataset adaptations.

\noindent Our contributions are summarized as follows:

\begin{itemize}
\item We recast blind 360° IQA as an active perception problem and propose RL-ScanIQA, the first reinforcement-learning-based framework that jointly learns scanpath generation and quality assessment directly from IQA supervision, without human scanpath annotations.

\item We develop a task-calibrated multi-level reward that converts sparse IQA supervision into dense shaping signals through step-wise exploration, set-level scanpath diversity, and task-aligned perceptual feedback.

\item We further employ distortion-space augmentation with rank-consistent training as auxiliary regularization to improve robustness under cross-dataset distribution shifts.

\item Extensive experiments on three 360° IQA benchmarks demonstrate that our proposed RL-ScanIQA achieves state-of-the-art performance in both in-dataset and cross-dataset evaluations. 
\end{itemize}

\section{Related Work}
\label{sec:related work}

\noindent\textbf{BIQA for Planar Images and Videos.} Early BIQA methods relied on natural scene statistics and handcrafted features \cite{mittal2012no,mittal2012making,liu2025difflow3d}. Recent works adopted convolutional and transformer architectures to predict perceptual quality from raw pixels~\cite{ke2021musiq,ma2025survey}. For planar videos \cite{ma2025ms,ma2025fine}, Korhonen et al.~\cite{korhonen2019two} and Li et al.~\cite{li2019quality} extended BIQA to handle temporal artifacts such as motion blur and frame drops. More recently, Li et al.~\cite{li2025q} proposed Q-Insight, a multi-modal reinforcement-learning framework for language-conditioned quality prediction. However, these approaches typically assume full-frame visibility and uniform attention across the frame, ignoring the spatial exploration behavior critical in immersive 360° environments~\cite{tu2021rapique}. This assumption limits their applicability to 360° content, where viewers perceive only a restricted viewport at any time.

\begin{figure*}
\includegraphics[width=\textwidth]{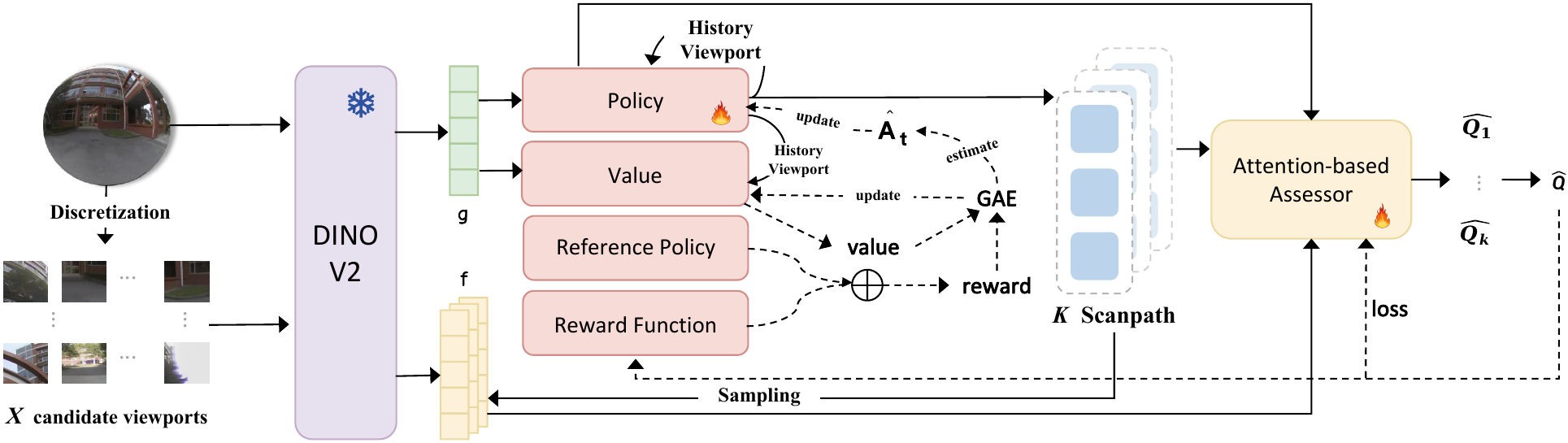}
\vspace{-6mm}
  \caption{Overview of RL-ScanIQA: the scanpath generator is trained with PPO policy, sampling K diverse scanpaths. The quality assessor evaluates each scanpath and produces the final quality score.}
  \vspace{-4mm}
  \label{framework}
\end{figure*}

\vspace{3pt}
\noindent\textbf{BIQA for 360° Images and Videos.} Unlike planar images \cite{liuarflow,ma2024generative}, 360° content restricts viewers to a limited viewport, making head and eye movements critical to quality perception. Early methods sampled fixed viewports or relied on equirectangular projection (ERP) to aggregate patch-level quality scores \cite{li2019viewport,li2018bridge,fu2022adaptive,yang2022tvformer}. Sun et al. \cite{sun2019mc360iqa} proposed MC360IQA, a multi-branch CNN processing six predefined viewports, while Xu et al. \cite{xu2020blind} introduced VGCN to model spatial relations among viewports via graph convolutions. However, these approaches relied on heuristic sampling and neglected sequential viewing behaviors. Recent methods have incorporated scanpath modeling to better capture user exploration. Wu et al. \cite{wu2023assessor360} employed recursive probability sampling to generate pseudo scanpaths and learned temporal dependencies through a distortion-aware network. Sui et al. \cite{sui2025perceptual} formulated scanpath generation as a generative modeling problem to simulate human-like viewing. Fan et al. \cite{fan2024learned} integrated scanpath predictions into a video-quality pipeline. However, these methods generate scanpaths via separate preprocessing or heuristic/imitation-based policies, that are decoupled from the IQA objective. Consequently, models cannot learn task-aligned exploration strategies.

\vspace{3pt}
\noindent\textbf{Reinforcement Learning for Visual Tasks.}
Reinforcement learning (RL) enables agents to learn decision-making policies through environmental interaction and reward maximization \cite{yang2024regulating}. Zhou et al. \cite{zhou2018deep} and Devrim et al. \cite{devrim2017reinforcement} applied RL to visual tasks involving sequential decisions, such as attention selection, video summarization, and viewpoint planning. Among RL algorithms, Proximal Policy Optimization (PPO) \cite{schulman2017proximal} has emerged as a robust on-policy method that stabilizes training via clipped surrogate objectives and entropy regularization. Stiennon et al. \cite{stiennon2020learning} demonstrated that PPO performed reliably with episodic rewards when combined with variance reduction and value bootstrapping. Motivated by these properties, we formulate the scanpath generation as a learnable policy and apply PPO to optimize the generated scanpath for perceptual quality assessment.


\section{Methodology}

\subsection{Overview}
As illustrated in Figure \ref{framework}, our method jointly optimizes a scanpath generator and a quality assessor within a reinforcement-learning-based framework. Given a 360° image, we first discretize the viewing sphere into $X$ candidate viewports and initialize the agent at a starting viewport. The RL policy generates $K$ scanpaths $S_k = {x_1^k, \dots, x_T^k}$, each consisting of $T$ viewports sampled from the $X$ candidates (section \ref{sec:scanpath}). The policy is guided by multi-level reward signals (section \ref{sec:reward}). The quality assessor then evaluates the image quality scores ${\hat{Q}_1, \dots, \hat{Q}_K}$ based on each scanpath, and the final image quality is estimated by averaging these scores (section \ref{sec:quality assessor}). Distortion-space data augmentation and rank-consistent losses are applied to enhance cross-domain performance (section \ref{sec:cross-domain}).


\subsection{Scanpath Generator}
\subsubsection{Sequential Scanpath Modeling} 
\label{sec:scanpath}
To capture the sequential and content-aware nature of human visual exploration, we formulate scanpath generation as a finite-horizon Markov Decision Process (MDP) \cite{sutton1998reinforcement,devrim2017reinforcement} with horizon length $T$. For each horizon step $t$, the agent operates over a discretized spherical action space comprising $\{x_{t}^j\}_{j=1}^X$, sampled uniformly on $X$ viewports covering the sphere. Each viewport is parameterized by its central yaw–pitch coordinate and has a fixed field of view (FOV). horizontal wrapping and polar clamping are applied for spherical continuity \cite{coors2018spherenet,chen2023panogrf}.

\vspace{3pt}
\noindent\textbf{Update of Agent State.} At each horizon step $t$, the policy conditions its decision on both historical trajectory and global scene context. Specifically, the agent state is defined as $s_t = [{h}_{t-1}; {g}]$, where ${h}_{t-1}$ is the hidden state of a Gated Recurrent Unit (GRU) \cite{cho2014learning} summarizing history viewports $\{x_1,\dots,x_{t-1}\}$, and ${g} \in \mathbb{R}^{1024}$ is a global descriptor extracted from a frozen DINOv2 encoder~\cite{oquab2023dinov2}. The next action is sampled from an auto-regressive policy $\pi_\theta(x_t \mid s_t)$, where $x_t$ denotes the selected viewport at time step $t$.


To enable content-aware decision making, the policy explicitly incorporates candidate viewport features. Given the precomputed viewport features $\{\mathbf{f}^j_{t}\}_{j=1}^X$ that are also extracted by the shared DINOv2 encoder, the next viewport is selected based on:
\begin{equation}
z^j_{t} = \mathbf{v}^\top \tanh\!\left(W_h \mathbf{h}_{t-1} + W_g \mathbf{g} + W_f \mathbf{f}^j_{t} + \mathbf{b}\right) + m_t(j),
\end{equation}
where $W_h \in \mathbb{R}^{d_z \times d_h}$, $W_g \in \mathbb{R}^{d_z \times 1024}$, and $W_f \in \mathbb{R}^{d_z \times 1024}$ are projection matrices; $\mathbf{b}, \mathbf{v} \in \mathbb{R}^{d_z}$ are learnable bias and scoring vectors; $m_t(j) \in \mathbb{R}$ is a dynamic mask applied to invalid viewports $j$ at step $t$. Consequently, the overall viewport selection at time $t$ can be represented by the policy $\pi_\theta(x_t \mid s_t) = \mathrm{Softmax}(\{z^j_t\}_{j=1}^X)$.

\vspace{3pt}
\noindent\textbf{Optimization Policy.} To train the auto-regressive policy $\pi_\theta$, we adopt Proximal Policy Optimization (PPO) \cite{schulman2017proximal}, a widely-used on-policy algorithm that balances stability and exploration. During each iteration, the agent collects experience from sampled scanpaths and computes return and advantage estimates using the multi-level rewards (section \ref{sec:reward}). The policy is then updated by minimizing the following clipped surrogate objective:
\begin{align}
\mathcal{L}(\theta) =\; 
&\mathbb{E} \left[ \min\left( \rho \hat{A}_t,\; \text{clip}(\rho, 1 - \epsilon, 1 + \epsilon) \hat{A}_t \right) \right] \notag \\
&\quad + c_v \cdot \mathbb{E} \left[ \left( V_{\phi}(s) - \mathcal{R}_\text{total} \right)^2 \right] \notag \\
&\quad - c_H \cdot \mathbb{E} \left[ \mathcal{H}\left( \pi_{\theta}(\cdot \mid s) \right) \right],
\end{align}
where $\rho = \pi_\theta(a \mid s) / \pi_{\theta_{\text{old}}}(a \mid s)$ is the importance ratio, $\hat{A}_t$ is the estimated advantage at time $t$, $c_v$ is the weight of the value loss, $V_\phi(s)$ is the value predicted by the critic network with parameters $\phi$, and $\mathcal{R}_\text{total}$ is the total discounted return computed from multi-level rewards (section \ref{sec:reward}). $H(\cdot)$ denotes the entropy of the current policy. 
The PPO clipping threshold $\epsilon$ and the entropy regularization coefficient $c_H$ are scheduled during training.

We estimate advantages $\hat{A}$ using Generalized Advantage Estimation (GAE) \cite{schulman2015high}. Given the critic $V_\phi(s_t)$ defined on the policy state $s_t=[h_{t-1};g]$, the temporal-difference residual and advantage are:
\begin{align}
\delta_t &= r_t + \gamma (1-d_t) V_\phi(s_{t+1}) - V_\phi(s_t),\\
\hat A_t &= \delta_t + \gamma\lambda (1-d_t)\hat A_{t+1},
\end{align}
where $\gamma$ and $\lambda$ denote the discount and trace parameters, respectively, and $d_t$ indicates episode termination.



\subsubsection{Design of Rewards} 
\label{sec:reward}
\begin{figure}
\includegraphics[width=\linewidth]{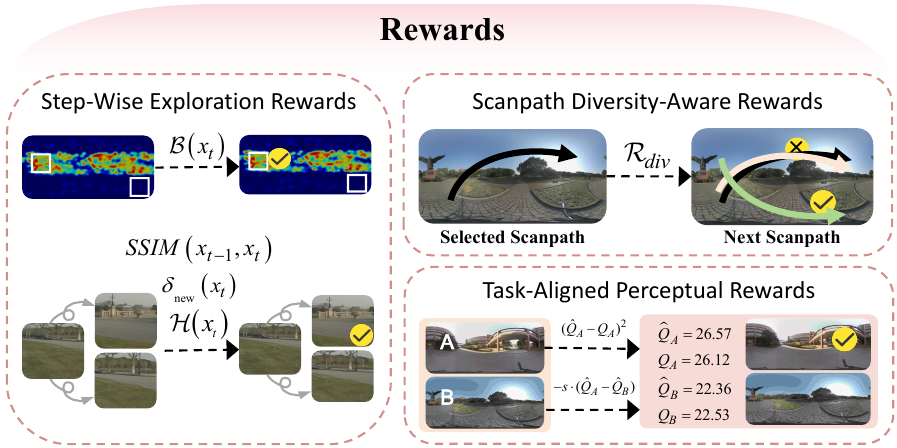}
\vspace{-7mm}
  \caption{Rewards: multi-level rewards are designed in our RL policy update. (More details and code in Supplementary Material)}
  \label{rewards}
  \vspace{-4mm}
\end{figure}

As shown in Figure \ref{rewards}, to guide the scanpath policy toward producing task-relevant, behaviorally informed viewing trajectories, we formulate a composite reward function from three cues: (1) Step-wise Exploration Rewards; (2) Scanpath Diversity-Aware Rewards; and (3) Task-Aligned Perceptual Rewards.

\vspace{3pt}
\noindent\textbf{A. Step-Wise Exploration Rewards.} To guide the agent toward generating task-relevant viewport sequences, we define a step-wise reward that encodes coarse viewing priors in immersive scenes. Rather than treating the sequence as a monolithic trajectory, we provide localized reward signals at each time step to enable fine-grained decision-making during navigation.

Specifically, at each step $t$, the agent receives a reward $r_t$ for the selected viewport $x_t$, which integrates four perceptual cues grounded in empirical findings. First, we incorporate $\mathcal{H}(x_t)$, the Shannon entropy of the viewport's grayscale histogram \cite{larkin2016reflections}, to encourage the agent to attend to structurally rich or texturally complex regions. These areas not only attract human gazes but also often contain high-frequency information that is sensitive to quality degradations \cite{renninger2004information, bruce2009saliency}. Second, to reduce redundancy and promote broader coverage, we include a dissimilarity term based on $1 - \text{SSIM}(x_{t-1}, x_t)$, which penalizes transitions to visually similar patches and steers the agent toward semantically diverse regions \cite{erdem2013visual,wang2019revisiting}. Third, we add a novelty signal $\delta_\text{new}(x_t)$ that discourages the revisiting of previously attended regions, enabling more comprehensive exploration of the spherical content. This is particularly important for detecting localized artifacts that may be missed during narrow or repetitive viewings \cite{itti2009bayesian, ghazizadeh2022salience}. Finally, motivated by eye-tracking studies on 360° content \cite{sitzmann2018saliency,xu2018gaze,jiang2020fantastic}, we introduce an equator-bias prior $\mathcal{B}(x_t)$ to softly favor viewports near the equator, where human attention tends to concentrate in panoramic scenes.

These components are linearly combined to form the final step-wise exploration rewards as:
\begin{align}
r_t =\;&
\lambda_{\text{ent}} \cdot \mathcal{H}(x_t)
+ \lambda_{\text{ssim}} \cdot \mathbf{1}[t>1] \cdot \left(1 - \mathrm{SSIM}(x_{t-1}, x_t)\right) \nonumber\\
&+ \lambda_{\text{nov}} \cdot \delta_{\text{new}}(x_t)
+ \lambda_{\text{eqb}} \cdot \mathcal{B}(x_t),
\end{align}
where $\mathbf{1}[\cdot]$ denotes the indicator function, $\delta_{\text{new}}(x_t)$ is assigned as 1 if $x_t \notin \{x_1,\dots,x_{t-1}\}$ and $0$ otherwise. \begin{equation}
B(x_t) = \exp\!\left(-\gamma_{\mathrm{eq}} \cdot \left| \mathrm{pitch}(x_t) \right|\right)
\end{equation}
softly prefers equatorial viewports, where $\gamma_{\mathrm{eq}}$ controls the equator-bias strength and $\mathrm{pitch}(x_t) \in [-\pi/2, \pi/2]$ denotes the latitude of $x_t$ in radians. The coefficients $\lambda_{\text{ent}}$, $\lambda_{\text{ssim}}$, $\lambda_{\text{nov}}$, and $\lambda_{\text{eqb}}$ balance the relative contributions of the four reward terms.

\vspace{3pt}
\noindent\textbf{B. Scanpath Diversity-Aware Rewards.} In panoramic scenes, perceived quality varies depending on which regions are attended. For example, compression artifacts may be more prominent in sky regions, while motion blur may occur near the ground \cite{xu2020viewport,kapoor2021capturing,sui2021perceptual}. To capture this variability and improve robustness to localized degradations, we introduce a set-level diversity reward. Moreover, since different human observers tend to focus on different regions of the same image, incorporating diverse scanpaths can help simulate this perceptual subjectivity and align better with averaged Mean Opinion Score (MOS) \cite{wu2023assessor360,fan2024learned,sui2025perceptual}. Given $K$ sampled scanpaths $\{S_1, \dots, S_K\}$ per image, we define a diversity bonus $\mathcal{R}_\text{div}$ as:
\begin{equation}
\mathcal{R}_\text{div} = \beta_\text{cov} \cdot \frac{|\cup_k S_k|}{X} - \beta_\text{jac} \cdot \frac{1}{K(K-1)} \sum_{i \neq j} \text{Jacc}(S_i, S_j),
\end{equation} 
where $\bigcup_k S_k$ represents the union of all visited viewports across $K$ scanpaths, and $\beta_{\text{cov}}$ and $\beta_{\text{jac}}$ control the coverage reward and overlap penalty, respectively. The first term rewards coverage, encouraging the union of all scanpaths to span a large portion of the viewing sphere. The second term penalizes overlap between scanpaths using pairwise Jaccard similarity \cite{liu2017finding,yang2020context}, ensuring that the policy generates a set of distinct and complementary scanpaths.

\vspace{3pt}
\noindent\textbf{C. Task-Aligned Perceptual Rewards.} Ultimately, the goal of scanpath generation is to support perceptual quality prediction. To enforce task alignment, we introduce two supervised rewards based on a context-aware evaluator network. Given a pair of images $(I_1, I_2)$ with associated ground truth MOS scores $(Q_1, Q_2)$ and predicted ones $(\hat{Q}_1, \hat{Q}_2)$ produced by the quality assessor (in Section \ref{sec:quality assessor}), we define the following two task-level rewards. First, a negative MSE reward $\mathcal{R}_\text{mse}$ penalizes the squared error between predicted and ground-truth scores for each image:
\begin{equation}
\mathcal{R}_\text{mse} = -\left[ (\hat{Q}_1 - Q_1)^2 + (\hat{Q}_2 - Q_2)^2 \right].
\end{equation} 
This encourages the scanpath policy to attend to regions that help the evaluator make accurate predictions.

Second, to capture relative quality perception, we introduce a pairwise ranking reward $\mathcal{R}_\text{rank}$ that promotes correct ordering between the two images: 
\begin{equation}
\mathcal{R}_\text{rank} = -\log\left(1 + \exp\left[-s \cdot (\hat{Q}_1 - \hat{Q}_2)\right]\right).
\end{equation}
where $s = \text{sign}(Q_1 - Q_2)$ denotes the ground‑truth ranking direction ($s{=}1$ if $Q_1>Q_2$, $-1$ otherwise$)$. This soft ranking formulation ensures smooth gradients even when score differences are small or ambiguous, and is especially useful when the perceptual quality gap between two distorted images is subtle.


The final reward for each image pair is obtained by aggregating all components described above. Specifically, for each sampled scanpath $S_k = \{x_1^k, \dots, x_T^k\}$, we compute the sum of its step-level exploration rewards. These are then combined with the set-level diversity reward and task-aligned supervision terms to form the total reward signal:
\begin{equation}
\mathcal{R}_\text{total} = \frac{1}{K} \sum_{k=1}^{K} \left( \sum_{t=1}^{T} r_t^{(k)} \right) + \mathcal{R}_\text{div} + \lambda_\text{mse} \cdot\mathcal{R}_\text{mse} + \lambda_\text{rank} \cdot\mathcal{R}_\text{rank},
\end{equation} 
where $\lambda_{\text{mse}}$ and $\lambda_{\text{rank}}$ are scheduled weighting factors for the two task-aligned supervision terms, and $r_t^{(k)}$ is the exploration reward at time step $t$ of scanpath $k$.


\subsection{Quality Assessor}

\subsubsection{Quality Prediction Module}
\label{sec:quality assessor}
The Quality Assessor (QA) module predicts the perceptual quality from a given scanpath. Each scanpath $S_{k}$ is rendered into rectilinear patches and encoded as feature vectors $\{\mathbf{f}_1, \dots, \mathbf{f}_T\}$ with DINOv2. The global image context $\mathbf{g}$ is similarly extracted by a shared DINOv2 encoder from the full ERP image as in Section \ref{sec:scanpath}.

To aggregate local viewport features in a context-aware manner, we adopt an attention-based pooling mechanism. Specifically, the attention weight $\alpha_t$ of each region-level feature $\mathbf{f}_t$ is computed by its interaction with the global feature $\mathbf{g}$, allowing the model to emphasize viewports that are more informative for quality assessment. The scanpath-level representation $\mathbf{m}_k= \sum_{t=1}^{T} \alpha_t \cdot \mathbf{f}_t^k$ is then obtained as a weighted sum:
\begin{equation}
\alpha_t = \frac{
\exp\left(\mathbf{v}^\top \tanh\left(\mathbf{W}_p \mathbf{f}_t^k + \mathbf{W}_g \mathbf{g} \right) \right)
}{
\sum_{i=1}^{T} \exp\left( \mathbf{v}^\top \tanh\left( \mathbf{W}_p \mathbf{f}_i^k + \mathbf{W}_g \mathbf{g} \right) \right)
}.
\end{equation}
where $\mathbf{f}_t^k$ is the feature of the $t$-th viewport in the $k$-th scanpath, $\mathbf{v} \in \mathbb{R}^{d_a}$ is a learnable attention vector, $\mathbf{W}_p$ and $\mathbf{W}_g$ are projection matrices for local and global features, respectively. The resulting representation $\mathbf{m}_k$ captures the semantics of the visited regions, modulated by their importance under the global image context. To incorporate global information, we concatenate $\mathbf{m}_k$ with the global feature $\mathbf{g}$ and feed the result into a multi-layer perceptron (MLP) to regress the scanpath-level quality score: $\hat{Q}_k = \mathrm{MLP}([\mathbf{m}_k; \mathbf{g}])$. This design enables the QA to predict a scalar perceptual score $\hat{Q}_k$ for each individual scanpath $S_k$, making it possible to model the quality implications of different viewing behaviors over the same image.


\subsubsection{Cross-Domain Enhancement} 
\label{sec:cross-domain}
\begin{figure}
\includegraphics[width=\linewidth]{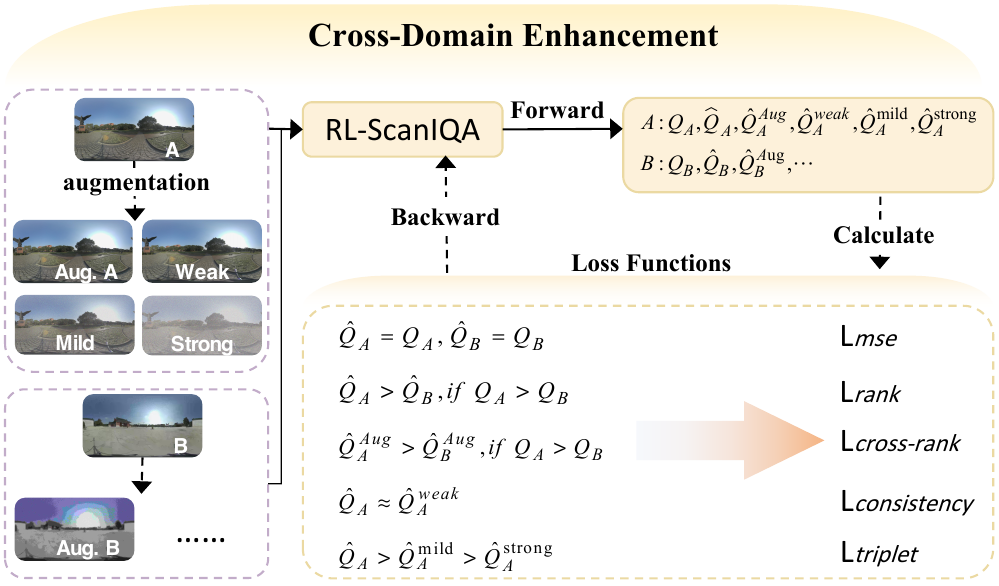}
\vspace{-7mm}
  \caption{Cross-Domain Enhancement: the quality assessor module is trained using multiple losses with distortion-space augmentation. (More details and code in Supplementary Material)}
  \label{losses}
  \vspace{-4mm}
\end{figure}

As shown in Figure \ref{losses}, to supervise the Quality Assessor (QA), we adopt two standard objectives on a pair of images $(I_1, I_2)$ with ground truth MOS scores $(Q_1, Q_2)$ and predicted ones $(\hat{Q}_1, \hat{Q}_2)$ from the quality assessor: a regression loss to align with ground-truth MOS scores $\mathcal{L}_\text{mse} = (\hat{Q}_1 - Q_1)^2 + (\hat{Q}_2 - Q_2)^2$ , and a pairwise ranking loss $\mathcal{L}_\text{rank} = \log\left(1 + \exp\left(-s \cdot (\hat{Q}_1 - \hat{Q}_2)\right)\right)$ to preserve image orderings. Beyond these losses, we also adopt three augmentation-based objectives to enhance robustness under distribution shifts: (1) Similarity Consistency Loss; (2) Triplet Loss; (3) Cross-Rank Loss.

\vspace{3pt}
\noindent\textbf{(A) Similarity Consistency Loss.}
To reflect the human insensitivity to small visual changes and mild distortions, we introduce a consistency loss that encourages the model to produce similar quality scores under weak perturbations. This regularization encourages smooth prediction behaviors. For each image, we generate a clean version and a weakly augmented version through compression, blur, or color jitter, and compute their respective predictions $\hat{Q}_\text{clean}$ and $\hat{Q}_\text{weak}$. The loss is defined as: $\mathcal{L}_\text{cons} = \big(\hat{Q}_\text{clean} - \hat{Q}_\text{weak}\big)^2$.

\vspace{3pt}
\noindent\textbf{(B) Triplet Loss.}
To enable the model to capture the diverse intensity of distortions, we further construct augmented triplets for each image: clean, mild, and strong. The predicted scores $\hat{Q}_\text{clean}$, $\hat{Q}_\text{mild}$, and $\hat{Q}_\text{strong}$ are expected to satisfy the ordering:
\begin{equation}
\begin{split}
\mathcal{L}_\text{triplet} =\ & \max(0, \hat{Q}_\text{mild} - \hat{Q}_\text{clean} + m_1) \\
& +\ \max(0, \hat{Q}_\text{strong} - \hat{Q}_\text{mild} + m_2) \\
& +\ \max(0, \hat{Q}_\text{strong} - \hat{Q}_\text{clean} + m_3),
\end{split}
\end{equation} 
where $m_1$, $m_2$, and $m_3$ denote the triplet margins.

\vspace{3pt}
\noindent\textbf{(C) Cross-Rank Loss.}
Finally, we ensure that relative quality judgments are preserved even after data augmentation. Given two images $I_A$ and $I_B$ with ground truth MOS $Q_A > Q_B$, we generate their mildly or strongly augmented versions and require that the ranking is still preserved. The loss is defined as:
\begin{equation}
\mathcal{L}_\text{cross} = \log\left(1 + \exp\left(-s \cdot (\hat{Q}_{A}^{\text{aug}} - \hat{Q}_{B}^{\text{aug}})\right)\right),
\end{equation} 
where $s = \text{sign}(Q_A - Q_B)$, and $\hat{Q}_{A}^{\text{aug}}$, $\hat{Q}_{B}^{\text{aug}}$ are the predictions after augmentation.

The final training loss of the QA head is composed by:
\begin{equation}
\begin{split}
\mathcal{L}_\text{total} =\ & \beta_\text{mse} \cdot \mathcal{L}_\text{mse}
+ \beta_\text{rank} \cdot \mathcal{L}_\text{rank}
+ \beta_\text{cons} \cdot \mathcal{L}_\text{cons} \\
& +\ \beta_\text{triplet} \cdot \mathcal{L}_\text{triplet}
+ \beta_\text{cross} \cdot \mathcal{L}_\text{cross}.
\end{split}
\end{equation} 
where $\beta_{\text{mse}}, \beta_{\text{rank}}, \beta_{\text{cons}}, \beta_{\text{triplet}},$ and $\beta_{\text{cross}}$ are loss weights.

\section{Experiments}
\subsection{Experimental Setups}
\subsubsection{Datasets}
We evaluate our method on three panoramic image IQA datasets: CVIQD\cite{sun2017cviqd}, OIQA\cite{duan2018perceptual}, and JUFE~\cite{fang2022perceptual}. CVIQD contains 528 compressed omnidirectional images derived from 16 references using JPEG, AVC, and HEVC codecs. OIQA includes 320 images derived from 16 references, distorted using JPEG, JPEG2000, Gaussian blur, and Gaussian noise. JUFE comprises 1032 distorted images derived from 258 references with non-uniform regional distortions that better reflect real-world degradations. The JUFE dataset also includes eye-movement data and provides four mean opinion scores (MOS) per image, collected using two starting points and two viewing durations (5s/15s). To be consistent with our single-label blind IQA setting, we average these four scores into one MOS per image. We also tested using only 15-second MOS (averaged over starting points) with similar results, but adopted the full-condition average for dataset consistency and fair baseline comparison.


\setlength{\tabcolsep}{0.4mm}
\begin{table}[t]
	\centering
	\footnotesize
    \caption{In-dataset experiment results. The methods in the first and second sections are full-reference and blind, respectively. Learning-based methods are retrained on the designated training split to ensure fair comparison. Handcrafted and LLM-based methods are applied directly to ERP images without training. Best results are in \textbf{bold}. The second best is \underline{underlined}.}
    \vspace{-7mm}
	\begin{center}
		{
        \resizebox{1.0\columnwidth}{!}
		{
			\begin{tabular}{l|l | c | c | c | c| c | c}
						\toprule
						&\multirow{2}{*}{\textbf{Method}} & \multicolumn{2}{c|}{\textbf{JUFE}} & \multicolumn{2}{c|}{\textbf{OIQA}} & \multicolumn{2}{c}{\textbf{CVIQD}} \\
\cline{3-8}
& &\textbf{SRCC} & \textbf{PLCC} & \textbf{SRCC} & \textbf{PLCC} & \textbf{SRCC} & \textbf{PLCC} \\
\hline
\noalign{\smallskip}
\multirow{8}{*}{\begin{tabular}[c]{@{}c@{}}\rotatebox{90}{Full-reference}\end{tabular}}&
SSIM \cite{wang2004image} & 0.027 & 0.011 & 0.298 & 0.314 & 0.668 & 0.705 \\
&VSI \cite{zhang2014vsi} & 0.045 & 0.053 & 0.398 & 0.422 & 0.921 & 0.935 \\
&WS-PSNR \cite{sun2017weighted} & 0.027 & 0.081 & 0.375 & 0.318 & 0.778 & 0.812 \\
&LPIPS \cite{zhang2018unreasonable} & 0.085 & 0.061 & 0.607 & 0.594 & 0.822 & 0.830 \\
&S-SSIM \cite{chen2018spherical} & 0.013 & 0.076 & 0.607 & 0.594 & 0.822 & 0.830 \\
&WS-SSIM \cite{zhou2018weighted} & 0.037 & 0.052 & 0.591 & 0.382 & 0.873 & 0.754 \\
&DISTS \cite{ding2020image} & 0.105 & 0.247 & 0.602 & 0.613 & 0.901 & 0.887 \\
\hline
\multirow{18}{*}{\begin{tabular}[c]{@{}c@{}}\rotatebox{90}{Blind}\end{tabular}}&NIQE~\cite{mittal2012making} & 0.552 & 0.592 & 0.745 & 0.736 & 0.893 & 0.872 \\
&BRISQUE~\cite{mittal2012no} & 0.705 & 0.642 & 0.915 & 0.914 & 0.905 & 0.922 \\
&DBCNN~\cite{zhang2018blind} & 0.512 & 0.443 & 0.811 & 0.882 & 0.805 & 0.873 \\
&MC360IQA~\cite{sun2019mc360iqa} & 0.502 & 0.623 & 0.875 & 0.906 & 0.877 & 0.892 \\
&VGCN~\cite{xu2020blind} & 0.495 & 0.473 & 0.902 & 0.938 & 0.946 & 0.932 \\
&MFILGN~\cite{zhou2021no} & 0.334 & 0.303 & 0.511 & 0.572 & 0.514 & 0.498 \\
&MP-BOIQA~\cite{jiang2021multi} & 0.311 & 0.423 & 0.778 & 0.804 & 0.885 & 0.902 \\
&TreS~\cite{golestaneh2022no} & 0.174 & 0.188 & 0.812 & 0.843 & 0.883 & 0.925 \\
&MANIQA~\cite{yang2022maniqa} & 0.073 & 0.125 & 0.431 & 0.322 & 0.511 & 0.587 \\
&AHGCN~\cite{fu2022adaptive} & 0.591 & 0.586 & 0.903 & 0.875 & 0.915 & 0.926 \\
&Clip-IQA~\cite{wang2023exploring} & 0.132 & 0.155 & 0.117 & 0.243 & 0.399 & 0.428 \\
&LIQE~\cite{zhang2023blind} & 0.023 & 0.039 & 0.577 & 0.681 & 0.723 & 0.844 \\
&Assessor360~\cite{wu2023assessor360} & 0.489 & 0.510 & \textbf{0.979} & \underline{0.945} & \underline{0.958} & \underline{0.963} \\
&F-VQA (A)~\cite{fan2024learned} & 0.725 & 0.821 & 0.891 & 0.905 & 0.953 & 0.904 \\
&GSR-X~\cite{sui2025perceptual} & \textbf{0.843} & \underline{0.857} & 0.922 & 0.937 & 0.805 & 0.957 \\
&VU-BOIQA~\cite{yan2025viewport} & 0.696 & 0.733 & 0.935 & 0.912 & 0.944 & 0.952 \\
&Q-Insight~\cite{li2025q} & 0.557 & 0.412 & 0.643 & 0.795 & 0.872 & 0.801 \\
\cline{2-8}
&\textbf{RL-ScanIQA} & \underline{0.816} & \textbf{0.902} & \underline{0.941} & \textbf{0.967} & \textbf{0.970} & \textbf{0.970} \\
						
						\bottomrule
						
				\end{tabular}
		}}
	\end{center}
    \label{tab.In-Dataset}
    \vspace{-4mm}
\end{table}

\subsubsection{Implementation Details}
We implement our method using PyTorch. The viewing sphere is discretized into $8 \times 4 = 32$ candidate viewports, each spanning a $90^\circ \times 90^\circ$ field-of-view (FOV) and rendered at a resolution of $224 \times 224$ \cite{fan2024learned,wu2023assessor360}. All viewports are projected from the equirectangular projection (ERP) image using bilinear interpolation. The scanpath generator uses 6 gated recurrent unit (GRU) modules. To ensure a fair comparison, the GRU policy selects the initial viewport solely based on global image features to avoid biasing the scanpath. We train the framework for 300 epochs using the Adam optimizer $(\beta_1 = 0.9, \beta_2 = 0.999)$, with learning rates of $3 \times 10^{-4}$ (policy) and $1 \times 10^{-4}$ (evaluator). The batch size is set to 4, and gradients are clipped to a maximum L2 norm of 1.0.

We conduct in-dataset and cross-dataset evaluations. For in-dataset evaluation, we randomly split each dataset (CVIQD, OIQA, JUFE) into 80\% training and 20\% testing. Following~\cite{wu2023assessor360,fan2024learned}, we repeat this random split 10 times and report average performance. For cross-dataset evaluation, we train the model on one of the two larger datasets (CVIQD or JUFE) and test it on the other two datasets. During inference, $K = 15$ and $T = 7$ steps, averaging quality predictions across all scanpaths. We use Spearman rank-ordered correlation (SRCC) and Pearson linear correlation (PLCC) as the evaluation metrics \cite{sui2025perceptual,wu2023assessor360,fan2024learned}.

\subsubsection{Compared Methods}
We compare our RL-ScanIQA against seven full-reference IQA methods: SSIM \cite{wang2004image}, VIF \cite{zhang2014vsi}, WS-PSNR \cite{sun2017weighted}, LPIPS \cite{zhang2018unreasonable}, S-SSIM \cite{chen2018spherical}, WS-SSIM \cite{zhou2018weighted}, DISTS \cite{ding2020image}. We also compare our RL-ScanIQA against 17 blind IQA methods for both 2D and 360° images, including DBCNN \cite{zhang2018blind}, MC360IQA~\cite{sun2019mc360iqa}, VGCN~\cite{xu2020blind}, MFILGN \cite{zhou2021no}, MP-BOIQA \cite{jiang2021multi}, TreS \cite{golestaneh2022no}, MANIQA \cite{yang2022maniqa}, AHGCN~\cite{fu2022adaptive}, Clip-IQA \cite{wang2023exploring}, LIQE \cite{zhang2023blind}, Assessor360~\cite{wu2023assessor360}, F-VQA (Assessor360)~\cite{fan2024learned}, GSR-X~\cite{sui2025perceptual}, VU-BOIQA \cite{yan2025viewport}, handcrafted methods (NIQE~\cite{mittal2012making}, BRISQUE~\cite{mittal2012no}), and a large-language model (Q-insight)~\cite{li2025q}.


\subsection{Experimental Results}

\setlength{\tabcolsep}{0.4mm}
\begin{table}[t]
	\centering
	\footnotesize
    \caption{Cross-dataset experiment results. Best results are in \textbf{bold}. The second best is \underline{underlined}.}
    \vspace{-7mm}
	\begin{center}
		{
        \resizebox{1.0\columnwidth}{!}
		{
			\begin{tabular}{l | cc | cc}
						\toprule
						\multirow{2}{*}{\textbf{Method}} 
& \multicolumn{2}{c|}{\makecell{Train: \textbf{CVIQD} \\ Test: \textbf{OIQA} / \textbf{JUFE}}} 
& \multicolumn{2}{c}{\makecell{Train: \textbf{JUFE} \\ Test: \textbf{CVIQD} / \textbf{OIQA}}} \\ 
\cline{2-5}
& \textbf{SRCC} & \textbf{PLCC} & \textbf{SRCC} & \textbf{PLCC} \\ 
\hline
\noalign{\smallskip}
MC360IQA~\cite{sun2019mc360iqa} & 0.765 / 0.432 & 0.827 / 0.615 & 0.275 / 0.317 & 0.311 / 0.459 \\
VGCN~\cite{xu2020blind} & 0.481 / 0.423 & 0.530 / 0.481 & 0.753 / 0.679 & 0.705 / \underline{0.741} \\
AHGCN~\cite{fu2022adaptive} & 0.501 / 0.443 & 0.695 / 0.527 & 0.601 / 0.711 & 0.611 / 0.498 \\
Assessor360~\cite{wu2023assessor360} & \underline{0.853} / 0.632 & \underline{0.887} / \underline{0.749} & 0.617 / 0.724 & 0.405 / 0.499 \\
GSR-X~\cite{sui2025perceptual} & 0.804 / \underline{0.765} & 0.831 / 0.694 & \textbf{0.782} / \underline{0.732} & \underline{0.733} / 0.611 \\
VU-BOIQA~\cite{yan2025viewport} & 0.533 / 0.517 & 0.624 / 0.610 & 0.503 / 0.432 & 0.415 / 0.490 \\
F-VQA (A)~\cite{fan2024learned} & 0.772 / 0.621 & 0.604 / 0.509 & 0.665 / 0.679 & 0.683 / 0.732 \\
\hline
\textbf{RL-ScanIQA} & \textbf{0.901} / \textbf{0.800} & \textbf{0.913} / \textbf{0.822} & \underline{0.771} / \textbf{0.755} & \textbf{0.802} / \textbf{0.833} \\
						
						\bottomrule
						
				\end{tabular}
		}}
	\end{center}
    \label{tab.Cross-Dataset}
    \vspace{-6mm}
\end{table}

\noindent\textbf{In-Dataset Evaluation.} As shown in Table \ref{tab.In-Dataset}, RL-ScanIQA achieves the highest PLCC across all three benchmarks and demonstrates state-of-the-art performance on CVIQD. Methods lacking scanpath modeling, such as NIQE and Q-Insight, exhibited significantly lower performance, highlighting the value of viewport-based evaluation that emulates the human perception of 360° content through limited field-of-view observations rather than full-sphere analysis. On JUFE, while GSR-X achieved a higher SRCC, our substantially higher PLCC indicates superior calibration with respect to absolute quality scores under authentic distortions, enabled by our adaptive viewport selection rather than uniform sampling. On OIQA, RL-ScanIQA's PLCC surpasses that of Assessor360 while maintaining competitive ranking ability. This gain results from jointly learning scanpaths with the quality assessor and the use of a learned prior, enabling the model to focus on distortion-relevant regions adaptively rather than relying on fixed viewing patterns.

\vspace{3pt}
\noindent\textbf{Cross-Dataset Evaluation.} 
As shown in Table \ref{tab.Cross-Dataset}, RL-ScanIQA exhibits strong cross-dataset generalization. Trained on CVIQD, our method achieves the highest PLCC on JUFE. Conversely, when trained on JUFE, we get the highest PLCC on both test sets. This robustness arises from the joint optimization of adaptive scanpaths with distortion-space augmentation and rank-consistency regularization, which produces more generalizable viewing strategies. In contrast to fixed-strategy methods, which exhibit significant performance declines across datasets, our learned policy appears to discover viewing patterns that generalize well across different distortion types and datasets by capturing the most salient features used to rate IQA.

\subsection{Qualitative Visualizations}

\begin{figure}
\includegraphics[width=\linewidth]{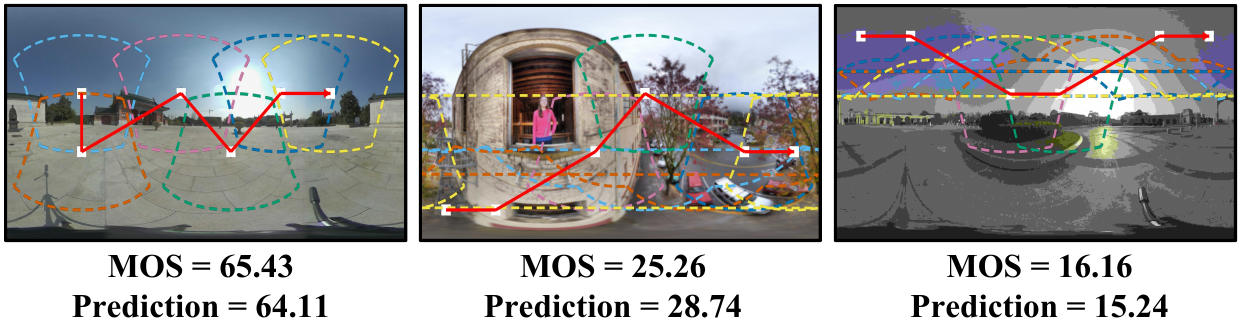}
\vspace{-6mm}
  \caption{Sampled discrete viewport sequences.  
  We visualize a representative scanpath (T=7) for each image and show the MOS and our predicted quality score below it.}
  \label{sampled viewports}
  \vspace{-2mm}
\end{figure}

\noindent\textbf{Sampled Viewports.}
In Figure \ref{sampled viewports}, we visualize three actual viewports selected by our policy. For these images, we illustrate one sampled discrete viewport sequence of length $T = 7$ by overlaying it on the ERP and displaying the corresponding rectilinear crops. For high-quality images, the policy exhibits broad and uniform coverage of salient content, while for low-quality images, it focuses on distortion-prone areas such as overcompressed sky and blurry ground. This visualization highlights the policy's ability to adapt its exploration strategy to image quality.

\begin{figure}
\includegraphics[width=\linewidth]{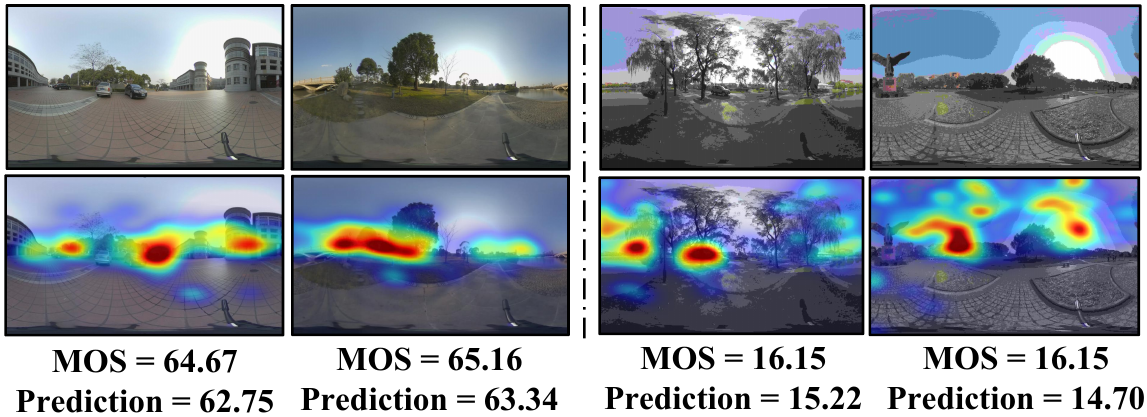}
\vspace{-6mm}
  \caption{Learned heatmaps: heatmaps aggregated from 15 scanpaths. \textit{Left}: High quality images (MOS=64.67, 65.16) show uniform exploration. \textit{Right}: Low quality images (MOS=16.15, 16.15) exhibit focused attention on degraded regions.
}
  \label{learned gaze}
  \vspace{-2mm}
\end{figure}

\vspace{3pt}
\noindent\textbf{Visualization of Attention.}
In Figure~\ref{learned gaze}, we visualized aggregated attention heatmaps generated by our learned scanpath policy for both high-quality (left) and low-quality (right) images. For high-quality content, the attention distribution exhibits balanced spatial coverage across salient regions. This reflects the exploratory behavior promoted by our diversity reward $\mathcal{R}_{\text{div}}$ and equator-bias prior $\mathcal{B}(r_t)$. In contrast, for low-quality images, attention concentrates on distortion-prone areas, such as the over-compressed sky. This indicates that our joint optimization allows the policy to identify task-relevant features and prioritize regions indicative of image quality.

\subsection{Ablation Studies}
\noindent\textbf{Scanpath Generator.} 
To assess the impact of the scanpath generator, we replace it with ground-truth human eye-tracking trajectories from the JUFE dataset~\cite{fang2022perceptual} (\emph{w/ Human GT Scanpaths}), where the quality assessor receives fixed human scanpaths as input. As shown in Table\ref{tab:ablation_jufe}, using human scanpaths results in a performance drop, which are attributed to two main factors: (1) the limited diversity of human scanpaths during training, which restricts the model's exposure to a range of viewing patterns, and (2) the semantic gap between unconstrained viewing patterns and the task-driven exploration required for effective image quality assessment. The human scanpaths may prioritize salient content rather than distortions critical for evaluating quality.

\setlength{\tabcolsep}{7mm}
\begin{table}[t]
\centering
\caption{Ablation study on the impact of scanpath generator and training paradigm on JUFE \cite{fang2022perceptual} dataset.}
\vspace{-3mm}
\label{tab:ablation_jufe}

\resizebox{1.0\columnwidth}{!}
		{
\begin{tabular}{lcc}
\toprule
\textbf{Method} & \textbf{SRCC} & \textbf{PLCC} \\
\midrule
w/ Human GT Scanpaths & 0.724 & 0.752 \\
w/o Joint Training & 0.651 & 0.783 \\
\textbf{Main Model} & \textbf{0.816} & \textbf{0.902} \\
\bottomrule
\vspace{-4mm}
\end{tabular}
}
\end{table}

\begin{figure}[t]
    \centering
    \includegraphics[width=\linewidth]{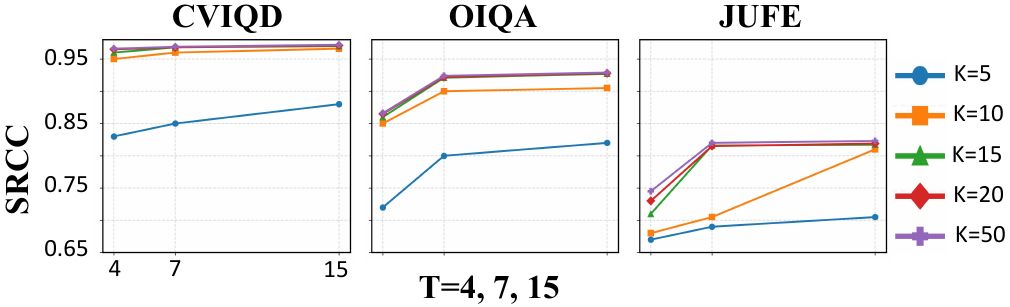}
    \vspace{-7mm}
    \caption{Ablation Study: Impact of the number of scanpaths (K) and the length of a scanpath (T).}
    \label{ablation}
    \vspace{-4mm}
\end{figure}

\vspace{3pt}
\noindent\textbf{Training Protocol.} 
We compare our joint training scheme with a two-stage separate training protocol (\emph{w/o Joint}), which involves: (1) training the scanpath policy via supervised imitation to match human eye-tracking trajectories using a cross-entropy loss, while keeping the visual feature extractor and the quality assessor fixed; and (2) freezing the weights of the learned policy and training the quality assessor independently using scanpaths generated by the now-fixed policy. The results in Table~\ref{tab:ablation_jufe} clearly indicate that joint optimization is critical to align the scanpath generation process with the IQA task objective, allowing the policy to uncover exploration strategies relevant to IQA.

\vspace{3pt}
\noindent\textbf{Number and Length of Scanpaths at Inference.} 
We explore whether the number $K$ and length $T$ of sampled scanpaths affect inference performance. We evaluate $K \in \{5,10,15,20,50\}$ and $T \in \{4,7,15\}$, averaging predictions across all sampled scanpaths. As shown in Figure \ref{ablation}, SRCC increases rapidly with larger $K$ and saturates around $K=15$, beyond which gains become marginal while inference cost grows linearly. For scanpath length, $T=7$ provides the best trade-off between accuracy and efficiency: (1) shorter sequences ($T=4$) suffer from insufficient exploration and lower performance, and (2) longer sequences ($T=15$) yield negligible improvements at considerably higher computational cost. 


\vspace{3pt}
\noindent\textbf{Reward Components.} To explore how each reward contributes to policy learning, we progressively remove them. We first remove all Step-wise Exploration Rewards, referred to as (\emph{w/o SER}). We then disable the Scanpath Diversity-aware Rewards (\emph{w/o SDR}), and finally remove the Task-aligned Perceptual Rewards (\emph{w/o TPR}). As shown in Table~\ref{tab:ablation_reward_compact}, the greatest degradation appears for \emph{w/o TPR} on CVIQD and OIQA, while \emph{w/o SDR} hurts JUFE the most. 


\setlength{\tabcolsep}{1mm}
\begin{table}[t]
\centering
\caption{Ablation study on multiple reward components.}
\vspace{-3mm}
\label{tab:ablation_reward_compact}
\resizebox{\columnwidth}{!}{
\begin{tabular}{lcccccc}
\toprule
\multirow{2}{*}{\textbf{Method}} & \multicolumn{2}{c}{\textbf{CVIQD}} & \multicolumn{2}{c}{\textbf{OIQA}} & \multicolumn{2}{c}{\textbf{JUFE}} \\
\cmidrule(lr){2-3} \cmidrule(lr){4-5} \cmidrule(lr){6-7}
& \textbf{SRCC} & \textbf{PLCC} & \textbf{SRCC} & \textbf{PLCC} & \textbf{SRCC} & \textbf{PLCC} \\
\midrule
w/o SER & 0.952 & 0.960 & 0.903 & 0.942 & 0.754 & 0.766 \\
w/o SDR & 0.946 & 0.958 & 0.897 & 0.946 & 0.731 & 0.755 \\
w/o TPR & 0.921 & 0.933 & 0.874 & 0.916 & 0.720 & 0.771 \\
\textbf{Main Model} & \textbf{0.968} & \textbf{0.977} & \textbf{0.941} & \textbf{0.967}& \textbf{0.816} & \textbf{0.902} \\
\bottomrule
\vspace{-4mm}
\end{tabular}
}
\end{table}

\setlength{\tabcolsep}{0.5mm}
\begin{table}[t]
	\centering
	\footnotesize
    \caption{Ablation study on distortion-space augmentation}
    \vspace{-7mm}
	\begin{center}
		{
        \resizebox{1.0\columnwidth}{!}
		{
			\begin{tabular}{l | cc | cc}
						\toprule
						\multirow{2}{*}{\textbf{Method}} 
& \multicolumn{2}{c|}{\makecell{Train: \textbf{CVIQD} \\ Test: \textbf{OIQA} / \textbf{JUFE}}} 
& \multicolumn{2}{c}{\makecell{Train: \textbf{JUFE} \\ Test: \textbf{CVIQD} / \textbf{OIQA}}} \\ 
\cline{2-5}
& \textbf{SRCC} & \textbf{PLCC} & \textbf{SRCC} & \textbf{PLCC} \\ 
\hline
 w/o Aug.
& 0.782 / 0.774 & 0.825 / 0.701 
& 0.664 / 0.703 & 0.663 / 0.702\\
\textbf{Main Model} 
& \textbf{0.901} / \textbf{0.800} & \textbf{0.913} / \textbf{0.822} 
& \textbf{0.771} / \textbf{0.755} & \textbf{0.802} / \textbf{0.833} \\				
						\bottomrule
						
				\end{tabular}
		}}
	\end{center}
    \label{tab:aug}
    \vspace{-3mm}
\end{table}

\vspace{3pt}
\noindent\textbf{Cross-Domain Enhancement.} To evaluate the effectiveness of distortion-space augmentation for cross-dataset generalization, we train a model variant (\textit{w/o Aug.}) that exclude augmentations such as JPEG compression, motion blur, defocus blur, color jitter, and Poisson noise. Consequently, we also remove all augmentation-related loss supervision, retaining only the regression loss $\mathcal{L}_{\text{mse}}$ and the standard ranking loss $\mathcal{L}_{\text{rank}}$. As shown in Table~\ref{tab:aug}, all performance metrics exhibited a rapid decline, proving that augmentations significantly improve generalization.

\section{Conclusion}

We proposed RL-ScanIQA, a reinforcement learning framework for BIQA, consisting of a scanpath generator and a quality assessor. The proposed scanpath generator formulates viewport selection as a sequential decision-making process that directly optimizes for the downstream IQA objective. For training and enhance robustness, we devise multi-level rewards and task-aligned supervision, complemented by distortion-space augmentation and rank-consistency constraints. 


\noindent\textbf{Acknowledgments}

\noindent This work was supported by Marsden Fund Council managed by
the Royal Society of New Zealand under Grant MFP-20-VUW-180.

{
    \small
    \bibliographystyle{ieeenat_fullname}
    \bibliography{main}
}


\end{document}